\documentclass[twoside,11pt]{article}

\usepackage{jmlr2e}

\usepackage{booktabs}
\usepackage{listings}
\usepackage{amsmath}
\usepackage{comment}
\usepackage{lineno,hyperref}
\usepackage{amssymb}
\usepackage{amsmath}
\usepackage{verbatim}
\usepackage[ruled,vlined]{algorithm2e}

\ShortHeadings{Glass-Box interactive Machine Learning with the Human-in-the-loop}{Holzinger et al.}
\firstpageno{1}

\begin{document}

\title{A glass-box interactive machine learning approach for solving NP-hard problems with the human-in-the-loop}


\author{\name Andreas Holzinger, Markus Plass, Katharina Holzinger \email a.holzinger@hci-kdd.org\\
       \addr Holzinger-Group, HCI-KDD, Institute for Medical Informatics/Statistics\\
       Medical University Graz, Austria
       \AND
       \name Gloria Cerasela Cri\c san \email ceraselacrisan@ub.ro \\
       \addr Faculty of Sciences\\
       Vasile Alecsandri University of Bac\v au, Romania
       \AND
       \name Camelia-M. Pintea \email dr.camelia.pintea@ieee.org  \\
       \addr Faculty of Sciences\\
       Technical University of Cluj-Napoca, Romania
       \AND
       \name Vasile Palade \email vasile.palade@coventry.ac.uk \\
       \addr Faculty of Engineering, Environment and Computing\\
       Coventry University, UK
       }


\maketitle

\begin{abstract}
The ultimate goal of the Machine Learning (ML) community is to develop algorithms that can \textit{automatically} learn from data, extract knowledge and to make decisions \textit{without any human intervention}. Such automatic Machine Learning (aML) approaches show impressive success, e.g. in speech recognition, recommender systems, autonomous vehicles, or image analysis. Recent results even demonstrate intriguingly that deep learning applied for automatic classification of skin lesions is on par with the performance of dermatologists, yet outperforms the average. As human perception is inherently limited to $\leq\mathbb{R}^3$ such approaches can discover patterns, e.g. that two objects are similar, in arbitrarily high-dimensional spaces what no human is able to do. Humans can deal simultaneously only with limited amounts of data, whilst "big data" is not only beneficial but necessary for aML. However, in health informatics, we are often confronted with only a small number of data sets or rare events, where aML suffer of insufficient training samples. Many problems are computationally hard, e.g. subspace clustering, k-anonymization, or protein folding. Here, interactive machine learning (iML) may be of help, where a human-in-the-loop contributes to reduce a huge search space through heuristic selection of suitable samples. This can reduce the complexity of $NP$-hard problems through the knowledge brought in by a human agent involved into the learning algorithm. A huge motivation for iML is that standard black-box approaches lack \textit{transparency,} hence do not foster trust and acceptance of ML among end-users. Most of all, rising legal and privacy aspects, e.g. with the new European General Data Protection Regulations (GDPR), make black-box approaches difficult to use, because they often are not able to explain \textit{why} a decision has been made, e.g. \textit{why} two objects are similar. All these reasons motivate to open the black-box to a glass-box.
In this paper, we present some experiments to demonstrate the effectiveness of the iML human-in-the-loop approach, particularly in opening the black-box to a glass-box and thus enabling a human directly to interact with an learning algorithm. We selected the Ant Colony Optimization (ACO) framework, and applied it on the Traveling Salesman Problem (TSP). The TSP-problem is a good example, because it is of high relevance for health informatics, e.g. for the study of protein folding, thus of enormous importance for fostering cancer research. Finally, from studies of learning from observation, i.e. of how humans extract so much from so little data, fundamental ML-research also may benefit.
\end{abstract}


\section{Introduction and Motivation for Research}

\textbf{Automatic Machine Learning.} The ultimate goal of the machine learning (ML) community is to develop algorithms/systems which can \textit{automatically} learn from data, extract knowledge and make predictions and decisions \textit{without any human intervention}~\citep{ShahriariAdamsFreitas:2016:HumanOutofTheLoop}. Such automatic machine Learning (aML) approaches have made enormous advance and practical success in many different application domains, e.g. in speech recognition~\citep{HintonEtAl:2012:DeepSpeech}, recommender systems~\citep{ChengEtAl:2016:DeepRecommenderSystems}, or autonomous vehicles~\citep{ChenEtAl:2015:DeepDriving}, and many other industrial applications.
Recently, deep learning algorithms, supported by supercomputers, cloud-CPUs and extremely large data sets \citep{DengLiEtAl:2009:ImageNet} have demonstrated to exceed human performance in visual tasks, particularly on playing games such as Atari \citep{MnihEtAl:2015:ReinforcementLearningNature}, or playing Go \citep{HassabisEtAl:2016:Go}.

An impressive example from the medical domain is the recent work by \citet{EstevaThrun:2017:DermaNN}: they utilized a GoogleNet Inception v3 CNN architecture \citep{SzegedyEtAl:2016:GoogleCNN} for the classification of skin lesions using a single CNN, trained end-to-end from images directly, using only pixels and disease labels as inputs. They pre-trained their network with 1.28 million images (1,000 object categories) from the 2014 ImageNet Challenge \citep{RussakovskyEtAl:2015:ImagenetChallenge}, and trained it on 129,450 clinical images, consisting of 2,032 different diseases. The performance was tested against 21 board-certified dermatologists on biopsy-proven clinical images with two critical binary classification use cases: keratinocyte carcinomas versus benign seborrheic keratoses; and malignant melanomas versus benign nevi. The results show that CNN achieves a performance on par with all tested human experts across both tasks, demonstrating the ability of classifying skin cancer by ML with a level of competence comparable to dermatologists. Consequently, \textit{automatic machine learning (aML)} works well when having large amounts of training data~\citep{SonnenburgSchoelkopf:2006:BigDataNecessity}, consequently ``big data'', which is often considered as burden - is here not only beneficial but necessary.

\textbf{Disadvantages of black-box approaches.} Besides of being resource intensive and data hungry, black-box approaches have, at least when applied to safety-critical domains such as the medical domain, one enormous drawback: black-box approaches lack transparency, i.e. they often are not able to explain \textit{why} a decision has been made. This does not foster trust and acceptance among end-users. Most of all, legal aspects make black-box approaches difficult: Under the new European General Data Protection Regulations (GDPR) taking effect on June, 1st, 2018, customers are given a right-to-be-forgotten \citep{MalleEtAl:2016:forgotten}, i.e. having their data deleted on request. Whilst ensuring privacy is both positive and necessary, this could lead to a competitive disadvantage for data-driven and data-dependent European companies. Consequently, it becomes important to understand what effects the removal of certain data will have on the performance of ML techniques. Moreover, the need for privacy-aware ML pipelines and the production of k-anonymized open data sets for real-world usage will become urgent \citep{MalleKieseHolzinger:2017:DoNotDisturb}.

\textbf{Representation Learning and Context.} The performance of ML algorithms is dependent on the choice of the \textit{data representations}. Current ML algorithms are still unable to \textit{extract the discriminative knowledge} from data. \citet{BengioVincent:2013:RepresentationLearning} argue that this can only be achieved if the algorithms can learn to identify and to \textit{disentangle the underlying exploratory factors} already existent among the low-level data. That entails that a truly intelligent algorithm is required to understand the \textit{context}, and to be able to discriminate between relevant and irrelevant features -- similarly as we humans can do. The questions \textit{"What is interesting?"} and \textit{"What is relevant?"} are inherently hard questions, and as long as we cannot achieve true intelligence with automatic approaches, we have to develop algorithms which can be applied by a human expert. Such an domain expert is likely to be aware of the context. Following the probabilistic perspective, this would mean that learning features from data can be seen as recovering a parsimonious set of latent random variables (i.e., according to \textsc{Occams's} razor, see \citep{Domingos:1999:OccamProblem} for a critical discussion), representing a distribution over the observed data to express a probabilistic model $p(x, h)$ over the joint space of the latent variables, $h$, and the observed data $x$. This approach fits well into the perspective of cognitive science \citep{WilsonDannLucasXing:2015:TheHumanKernel}.


\textbf{Motivation for a human-in-the-loop.} Interactive Machine Learning (iML) has various definitions and is often used for any user-facing machine learning approach \citep{AmershiEtAl:2014:iML}. Other definitions speak also of a human-in-the-loop, but it is what we would rather call classic supervised ML approaches \citep{ShyuEtAl1999PhysicianInTheLoopCOMPVIS}, and a total different meaning is to put the human into physical feedback loops \citep{SchirnerEtAl:2013:hilCyberPhysical}.

By integrating a human-in-the-loop (e.g., a human kernel \citep{WilsonDannLucasXing:2015:TheHumanKernel}), or the involvement of a human directly into the algorithm, iML is defined by \textit{"algorithms which interact with agents and can optimize their learning behaviour through this interaction -- where the agents can be humans \citep{Holzinger:2016:definitionIML}".} The general idea is making use of the strengths of human cognitive abilities - when automatic approaches fail. Consequently, iML-approaches can be of particular interest to solve problems, where we are lacking big data sets, deal with complex data and/or rare events, where aML suffer of insufficient training samples.

In the medical domain a “doctor-in-the-loop” can help with his/her expertise in solving problems which otherwise would remain $NP$-hard. A recent experimental work \citep{HolzingerEtAl:2016:iMLExperiment} demonstrates the usefulness on the Traveling Salesman Problem (TSP), which appears in a number of practical problems, e.g., the native folded three-dimensional conformation of a protein in its lowest free energy state; or both 2D and 3D folding processes as a free energy minimization problem belong to a large set of computational problems, assumed to be conditionally intractable \citep{CrescenziEtAl:1998:ProteinFolding}. As the TSP is about finding the shortest path through a set of points, it is an intransigent mathematical problem, where many heuristics have been developed in the past to find approximate solutions \citep{MacgregorOrmerod:1996:HumanTSP}. There is evidence that the inclusion of a human can be useful in numerous other problems in different application domains, see e.g., \cite{NapolitanoEtAl:2008:ClusteringCell,AmatoEtAl:2006:MultiStepGene}. However, for clarification, iML means the integration of a human into the \textit{algorithmic} loop, i.e., to open the black box approach to a glass box. Other definitions speak also of a human-in-the-loop, but it is what we would call classic supervised approaches \citep{ShyuEtAl1999PhysicianInTheLoopCOMPVIS}, or in a total different meaning to put the human into physical feedback loops \citep{SchirnerEtAl:2013:hilCyberPhysical}.

The aML-approaches sometimes fail or deliver unsatisfactory results, when being confronted with complex problem. Here, interactive Machine Learning (iML) may be of help and a \textit{``human-in-the-loop"} may be beneficial in solving computationally hard problems, where human expertise can help to reduce, through heuristics, an exponential search space.

\begin{figure*}[ht]
	\centering
	\includegraphics[scale=0.225]{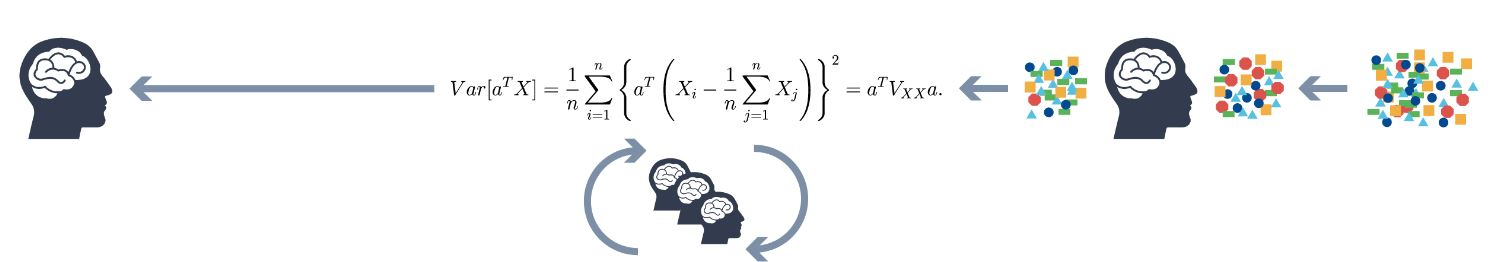}
	\caption{The iML human-in-the-loop approach: The main issue is that humans are not only involved in pre-processing, by selecting data or features, but actually during the learning phase they are directly interacting with the algorithm, thus shifting away the black-box approach to a glass-box; there might also be more than one human agent interacting with the computational agent(s), allowing for crowdsourcing or gamification approaches \citep{Holzinger:2016:NIPSiML}}.
	\label{fig:human-in-the-loop}
\end{figure*}

We define iML-approaches as algorithms that can interact with \textit{both computational agents and human agents} and can optimize their learning behaviour through these interactions \citep{Holzinger:2016:definitionIML}. To clearly distinguish the iML-approach from a classic supervised learning approach, the first question is to define the human's role in this loop (see Figure \ref{fig:human-in-the-loop}), \citep{Holzinger:2016:human-in-the-loop}.

\smallskip

There is evidence that humans sometimes still outperform ML-algorithms, e.g. in the instinctive, often almost instantaneous interpretation of complex patterns, for example, in diagnostic radiologic imaging: A promising technique to fill the semantic gap is to adopt an expert-in-the-loop approach, by integrating the physician’s high-level expert knowledge into the retrieval process and by acquiring his/her relevance judgments regarding a set of initial retrieval results \citep{AkgulBurakEtAl:2011:cbirExpertInTheLoop}.

Despite these apparent assumption, so far there is little quantitative evidence on the effectiveness and efficiency of iML-algorithms. Moreover, there is practically no evidence of \textit{how} such interaction may really optimize these algorithms, as it is a subject that is still being studied by cognitive scientists for quite a while and even though "natural" intelligent agents are present in large numbers throughout the world \citep{GigerenzerGaissmaier:2011:HeuristicDecisionMaking}.

From the theory of human problem solving, it is known that for example, medical doctors can often make diagnoses with great reliability - but without being able to explain their rules explicitly. Here iML could help to equip algorithms with such ``instinctive" knowledge and learn thereof. The importance of iML becomes also apparent when the use of automated solutions due to the incompleteness of ontologies is difficult \citep{PuppeEtAl:2006:InteractiveKnowledgeRefinement}.

This is important as many problems in machine learning and health informatics are ${\cal NP}$-hard, and the theory of ${\cal NP}$-completeness has crushed previous hopes that ${\cal NP}$-{\it hard} problems can be solved within polynomial time \citep{Papadimitriou:2003:complexity}. Moreover, in the health domain it is sometimes better to have an approximate solution to a complex problem, than a perfect solution to a simplified problem - and this within a reasonable time. Time is critical, as a clinical doctor in the routine has on average less than five minutes to make a decision \citep{Gigerenzer:2008:GutFeelings}.

Consequently, there is much interest in approximation and heuristic algorithms that can find near optimal solutions within reasonable time. Heuristic algorithms are typically among the best strategies in terms of efficiency and solution quality for problems of realistic size and complexity. Meta-heuristic algorithms are widely recognized as one of the most practical approaches for combinatorial optimization problems. Some of the most useful meta-heuristic algorithms include genetic algorithms, simulated annealing and Ant Colony Optimization (ACO).

\textbf{Learning from very few examples.} A further motivation for our research is the fact that humans are often surprisingly good in learning from very few examples. A good example are Gaussian processes, where aML approaches (e.g., kernel machines \citep{HofmannSchoelkopfSmola:2008:KernelMethods}) struggle on function extrapolation problems, which are quite trivial for human learners \citep{Griffiths:2009:HumanFunctionGaussian}.


As one of the {\it NAE Grand 21st Century} challenges is {\it health informatics}, this paper is a step towards the knowledge enhancement of this effervescent research domain \footnote{National Academy of Engineering Grand Challenges Health Informatics, 2015.\\
URL http://www.engineeringchallenges.org/challenges/informatics.aspx}.

Future information systems will have to address the balance between personal data privacy and seamless data sharing. Ubiquitous computing will improve health care and will pose new challenges to the networking systems. Future robust computerized systems will survey and react to public health emergencies both at local and global scale.

In this paper we address the question whether and to what extent a human can be beneficial in direct interaction with an algorithm. For this purpose we developed a Web-platform in the context of our iML-project \footnote{http://hci-kdd.org/project/iml (click on experiment) last accessed: 30.07.2017, 17:00 CEST}, to evaluate the effectiveness of the "human-in-the-loop" approach and to discuss some strengths and weaknesses of humans versus computers. Such an integration of a "human-into-the-loop" may have many practical applications, as e.g. in health informatics, the inclusion of a ``doctor-into-the-loop" \citep{Kieseberg:2016:DiL}, \citep{KiesebergEtAl:2016:ERCIMnews} can play a significant role in support of solving hard problems, particularly in combination with a large number of human agents (crowdsourcing, and collaborative interactive ML \cite{RobertBuettnerRoeckerHolzinger:2016:CollaborativeMachineLearning}  ).

\section{Background and Related Work}

\subsection{Problem Solving: Human versus Computer}

Many ML approaches perform very badly on extrapolation problems. This is mostly due to the fact that rarely one has exact function values without any errors or noise, which is therefore not suitable for solving real-world problems~\citep{AuerEtAl:1995:FunctionLearning}. Interestingly, such problems are very easy for humans, e.g. shown in a recent experiment by~\cite{WilsonDannLucasXing:2015:TheHumanKernel}: $N=40$ humans were presented functions drawn from Gaussian Processes (GP) with known kernels in sequence \citep{WilsonAdams:2013:GaussianProcessKernels}. A kernel function measures the similarity between two data objects. Formally, a kernel function takes two data objects $x_{i}$ and $x_{j} \in \mathbb{R}^{d}$, and produces a score $K: \mathbb{R}^{d} \times \mathbb{R}^{d} \rightarrow \mathbb{R}$.
Such a function can also be provided by a human to the machine learning algorithm, thus it is called: \textit{human kernel.} Automatically, this is done by the Support Vector Machine (SVM), because under certain conditions a kernel can be represented by a dot product in a high-dimensional space (Mercer's theorem) \citep{SteinwartScovel:2012:Mercer}. One issue here is that a kernel measures the similarity between two data objects, however, it cannot explain \textit{why} they are similar - here a human-in-the-loop can be of help to find the underlying explanatory factors of why two objects are similar, which requires context understanding in the domain.


In the experiment by~\cite{WilsonDannLucasXing:2015:TheHumanKernel} the human learners were asked to make extrapolations on these problems in sequence, so having an opportunity to progressively learn about the underlying kernel in each set. Consequently, they could repeat the first function at the end of the experiment, for six functions in each set; then they were asked for extrapolation judgments, because it provides more information about inductive biases than interpolation and pose difficulties for conventional GP kernels. The open question is still: how do humans do that. Even little children can learn surprisingly good from very few examples and grasp the meaning, because of their ability to generalize and apply the learned knowledge to new situations \citep{XuTenenbaum:2007:wordLearningBayesian}. The astonishing good human ability for inductive reasoning and \textit{concept generalization from very few examples} could derive from a prior combined with Bayesian inference \citep{TenenbaumEtAl:2011:GrowMind}: Let $X=x_1, ..., x_n$ be a set of $n$ observations from a novel entity $C$ (e.g. a word) and the examples given are from known entities $U$. It is assumed that the human learner has access to a hypothesis space $\mathcal{H}=\{{H_1}, {H_2}, ... , {H_n}\}$, and a probabilistic model relating each hypothesis $h in \mathcal{H}$ to data $X$. Simplified, each $h$ can be seen as a pointer to a subset of entities in the domain that is a candidate extension for $C$. The learner is assumed to be able to identify the extension of each $h$. Given the examples $X$, the Bayesian learner evaluates all hypotheses for candidate meanings according to Bayes’ rule, by computing their posterior probabilities $p(h|X)$, proportional to the product of prior probabilities $p(h)$ and likelihoods $p(X|h)$:

\begin{equation}\label{eq:a1}
p(h | X) = \frac{p(X | h)p(h)} {\sum_{\hslash \in \mathcal{H}} p(X|\hslash) p(\hslash)}\propto p(X | h) p(h).
\end{equation}

The prior $p(h)$ represents the learner’s expectation about plausible meanings for entity $C$ and is independent of the observed examples $X$ and reflects conceptual constraints, different contexts, or beliefs conditional on the meanings of other previously learned entities.

The likelihood $p(X|h)$ captures the statistical information inherent in the examples $X$, and reflects expectations about which entities are likely to be observed as examples of $C$ given a particular hypothesis $h$ about the meaning of $C$.

The posterior $p(h|X)$ reflects the learner’s degree of belief that $h$ is in fact the true meaning of $C$, given a combination of the observations $X$ with prior knowledge about plausible word meanings.

Whilst such a Bayesian approach provides a powerful computational framework for explaining how humans solve inductive problems, much open questions are remaining and we are still far away from being able to give answers on the question of how humans can gain out so much from so little data \citep{ThakerTenenbaumGershman:2017:OnlineSymbolic}. More background on these problems can be found e.g. in \citep{ChaterTenenbaumYuille:2006:ModelsOfCognition}, \citep{SteyversTenenbaumEtAl:2003:InferringCausalNetworks}.



\textbf{Why is this important for machine learning?} A true computationally intelligent algorithm should be able to automatically learn from data, extract knowledge and make decisions without any human intervention. Consequently machine learning efforts have always been inspired by \textit{how} humans learn, extract knowledge and make decisions. Key insights from past research provided probabilistic modelling and neurally inspired algorithms (see e.g. \cite{WolpertGhahramaniJordan:1995:Science}.  
From a probabilistic perspective, the ability for a model to automatically discover patterns and perform extrapolations is determined by a priori possible solutions and a priori likely solutions. Such a model should represent many possible solutions to a given problem with inductive biases, which can extract intricate structures from limited data.

In this context function learning is central for everyday cognition: nearly every task requires the construction of mental representations that map inputs to outputs $f: X \rightarrow Y$. Since the space of such mappings is infinite, inductive biases are necessary to constrain plausible inferences. Theories of how humans learn such relationships between continuous variables have focused on two possibilities: 1) humans are just estimating explicit functions, or 2) humans are performing associative learning supported by similarity principles. \citet{LucasGriffithsWilliamsKalish:2015:functionLearning} developed a rational model of human function learning that combines the strengths of both these theories.


\textbf{Human Abilities on Optimization Problems.} Getting insight into human abilities involved in optimization problems is considered of raising interest in the next decade~\citep{MacGregorChu:2011:HumanPerformanceTSP} as it could increase our knowledge about algorithms implemented by the visual system. The involved, relatively low complexity could be utilized to help to solve several optimization problems. Most of all, it could help to solve hard problems, because many computational tasks of practical interest and of relevance to the health informatics domain are exceptionally difficult to solve~\citep{MonassonEtAl:1999:ComplexityNature}.

\citet{MacgregorOrmerod:1996:HumanTSP} compared human performance on the Traveling Salesman Problem (TSP) to heuristic algorithms including Nearest Neighbor, Largest Interior Angle and Convex Hull algorithms. They concluded that these heuristics provided little to explain human performance, and that human problem solvers instead used a \textbf{perceptual process} to solve the problem; this was confirmed by two experiments carried out by \citet{BestSimon:2000:HumanPerformanceTSP}: They carried out two experiments to collect human solution data by a computer including mouse moves and clicks. Their first experiment replicated the problem set used by \citet{MacgregorOrmerod:1996:HumanTSP}. The second experiment used problems with uniform random distributions of points. Mouse movements were examined at the level of selection of individual moves (clicks). The TSPs were presented one at a time on a computer screen. Each problem started with a dialog box centered on the screen asking whether the participant was ready to start, and by clicking the “OK” button on the dialog, they centered their mouse. The participants then used the mouse to select nodes until completing a tour. They could not backtrack or undo moves. The results indicated that the Nearest Neighbor algorithm did explain a very large portion of human TSP solution methods. Although the overall solution obtained using a Nearest Neighbor algorithm did not produce solutions of the same quality as human problem solvers, a majority of the moves made by human problem solvers are to the closest point. When the local processing constraints suggested by the Nearest Neighbor algorithm are combined with a global plan for the general shape of the solution, the fit of the simulation to human performance was quite close.


There is evidence that humans are exceptionally good in bringing up near optimal solutions in decision making under uncertainty generally, and to the TSP specifically. Moreover, it is interesting that humans are not even realizing how hard such problems would be to solve computationally, but are able to exploit structural properties of the overall instance to improve parts of it \citep{KnillPouget:2004:BayesianBrain}, \citep{TenenbaumGriffithsKemp:2006:BayesianInductive}.

\citet{AcunaParada:2010:PeopleTSP} tested the solutions of N=28 participants to M=28 instances of the Euclidean Traveling Salesman Problem \citep{VanRooijEtAl:2003:EuclideanTSP}. In their experiment the participants were provided feedback on the cost of their solutions and were allowed unlimited solution attempts (as many trials as they liked). The authors found a significant improvement between first and last trials, and that solutions were significantly different from random tours that follow the convex hull and do not have self-crossings. They also found that participants modified their current better solutions so, that edges belonging to the optimal solution (the good edges) were significantly more likely to stay than other edges (the bad edges), which is an indication for good structural exploitation. More trials harmed the participants’ ability to tell good from bad edges, suggesting that after too many trials the participants simply ran out of ideas.

\noindent Research in this area, i.e. at the intersection of cognitive science and computational science is fruitful for further improving automatic aML approaches thus improving performance on a wide range of tasks, including settings which are difficult for humans to process (for example big data and high dimensional problems); on the other hand such experiments may provide insight into brain informatics.

\textbf{Human vs. Computer.} As mentioned in the introduction, aML algorithms outperform humans for example in high-dimensional data processing, in rule-based environments, or in automatic processing of large quantities of data (e.g. image optimization). However, aML-algorithms have enormous problems when lacking contextual information, e.g. in natural language translation/curation, or in solving ${\cal NP}$-{\it hard} problems. One important issue is in so-called \textit{unstructured problem solving:} Without a pre-set of rules, a machine has trouble solving the problem, because it lacks the \textit{creativity} required for complex problem solving. A good example for the literal competition of the human mind and its supposed artificial counterpart are various games, because they require human players to use their skill in logic, strategic thinking, calculating and/or creativity. Consequently, it is a good method to experiment on the strength and weaknesses of both humans and algorithms. Actually, the field of ML started with such efforts: In 1958 the first two programs to put the above question to the test were a checker program by \citep{Samuel:1959:machineLearningCheckers} and the first full chess program by Alex Bernstein~\citep{bernstein1958chess}. Whilst Samuel's program managed to beat Robert Nealey, the Connecticut checkers champion at that time, chess proved to be the computers weakness at that time; because on average just one move offers a choice of 30 possibilities, with an average length of 40 moves that leaves $10^{120}$ possible games. Recently, computers had impressive results in competitions against humans: In 1997, the world chess champion Gary Kasparov lost a six-game match against Deep Blue. A more recent example is the 2016 Google DeepMind Challenge, a five-game match between the world Go champion Lee Sedol and AlphaGo, developed by the Google DeepMind team. Although AlphaGo won the overall game, it should be mentioned that Lee Sedol won one game. This remark just should emphasize how much potential a combination of both sides may offer \citep{Holzinger:2013:HCI-KDD}.

As a test case for our approach, we selected the Traveling Salesman Problem, which is a classical hard problem in computer science and of high relevance for the health domain. The TSP has been studied for a long time, and Ant Colony Optimization has been used to provide approximate solutions \citep{cricsan2016ant}.

\subsection{Traveling Salesman Problem (TSP)}

The {\it Traveling Salesman Problem} is a {\it Combinatorial Optimization Problem} aiming to find the shortest cycle that passes through all the vertices of a complete weighted graph exactly once. {\it TSP} is one of the most studied problems in {\it Operational Research}, both for theoretical and practical reasons. With roots in the second half of the 18th century~\citep{laporte2006short}, it has now many variants, particularizations, generalizations, solving approaches and applications~\citep{applegate2011traveling}. {\it TSP} has strong connections with computational biology - construction of the evolutionary trees~\citep{korostensky2000using}, genetics - {\it DNA} sequencing~\citep{karp1993mapping}, designing of healthcare tools~\citep{nelson2007modeling}, or healthcare robotics~\citep{kirn2002ubiquitous}.

One of the most known definitions of the {\it TSP} is~\citep{applegate2011traveling}:

\noindent Definition 1. On a complete graph $G = (V, E)$, where $V$ is a set with $n$ vertices, let $(c_{ij})_{1\leq i,j\leq n}$ being the cost matrix associated with $E$. The goal of {\it TSP} is to find a minimum cost Hamiltonian circuit (i.e. a minimum cost path that passes through any vertex once and only once).

	From the computational complexity point of view, {\it TSP} is a $NP$-hard problem, meaning that it is very unlikely that a polynomial time algorithm for solving the worst case exists~\citep{garey1979computers}.

As an integer linear program, {\it TSP} supposes that the vertices form the set $\{1, 2,\dots, n\}$, defines the integer variables $(x_{ij})_{1\leq i,j\leq n}$  which describe a path:

\begin{equation}\label{eq:c1}
 x_{ij}=\left\{ \,
\begin{array}{cc}
   1  & \mbox{if the edge $(ij)$ is used}\\
  	0  & \mbox{otherwise}
\end{array}\right.
\end{equation}

\noindent and has the objective ~(\ref{eq:c2}) subject to the constraints ~(\ref{eq:c3})-~(\ref{eq:c5})~\citep{dantzig1959truck}.

\begin{equation}\label{eq:c2}
min(\sum^{n}_{i,j=1}x_{ij}c_{ij}).
\end{equation}

\begin{equation}\label{eq:c3}
  \sum^{n}_{i=1}x_{ij}=1 \hspace{0.5cm} \forall \hspace{0.2cm}  1 \leq j \leq n,
\end{equation}

\begin{equation}\label{eq:c4}
\sum^{n}_{j=1}x_{ij}=1 \hspace{0.5cm}\forall \hspace{0.2cm} 1 \leq i \leq n,
\end{equation}

\begin{equation}\label{eq:c5}
\sum_{i,j\in S}x_{ij} \leq | S | - 1  \hspace{0.2cm} \forall S \subset V,  \hspace{0.2cm} 2 \leq | S |\leq n-1.
\end{equation}

The constraints~(\ref{eq:c3}) and~(\ref{eq:c4}) ensure that only two selected edges are incident to each vertex. The constraint~(\ref{eq:c5})eliminates the subtours (the cycles using only the vertices from a proper subset $S$ of $V$).

Particularized {\it TSP} consider supplementary constraints. In the symmetric {\it TSP} case, the cost matrix is symmetric. It this is not true, the {\it TSP} is asymmetric. Metric {\it TSP} means that the costs form a metric that obeys the triangle inequality (in this case, the costs on edges can be seen as distances). {\it Euclidean TSP} is a metric {\it TSP} with the distances computed using the {\it Euclidean} norm.

{\it TSP} variants usually describe {\it TSP}-equivalent problems. For example, the clustered {\it TSP} partitions the vertices set $V$ into $k$ clusters and imposes that each cluster is visited in full after passing to next cluster. This problem is {\it TSP}-equivalent, as a large value can be added to the cost of any inter-cluster edge~\citep{chisman1975clustered}.

{\it TSP} generalizations usually model more complex real-life situations. Time dependent {\it TSP} has many formulations, all introducing variable values stored in the cost matrix $C$~\citep{gouveia1995classification}. {\it Vehicle Routing Problem (VRP)} asks for one optimum set of routes for an available set of vehicles that has to visit a given set of clients~\citep{dantzig1959truck}.

The most known {\it TSP} instance collections are \footnote{https://www.iwr.uni-heidelberg.de/groups/comopt/software/TSPLIB95/} and \footnote{http://www.math.uwaterloo.ca/tsp/data/index.html}. They serve as benchmarks for the implementations of the new methodologies for approaching TSP and its connected problems. Modern features as geographic coordinates can be included in the data~\citep{cricsan2017emergency}.

The computational complexity of the {\it TSP} makes that for large instances the exact approaches become impractical. This is why the heuristic methods are largely investigated and their solutions are widely accepted. The {\it Ant Colony Optimization}, initially designed for solving the {\it TSP}, is presented in the next Section.

\section{Ant Colony Optimization (ACO)}

The {\it Ant Colony Optimization (ACO) } is a metaheuristic inspired by the real ants’ ability to quickly find short paths from food to nest. As one of the most successful swarm/ social based livings on the planet, ants are able to form complex social systems. Without central coordination and external guidance, the ant colony finds the most efficient path between two points based on indirect communication. When moving, an ant deposits on the ground a chemical substance, called pheromone. The ants that arrive later, detect the trace and more likely follow it. Current investigations show more complex and elaborated behavior for specific species~\citep{franks2009speed}.

	The artificial ant colonies are multi-agent systems that work on problems represented by a graph. After each agent concurrently constructs a path in the graph, the pheromone is deposited on traversed edges and the process is iterated until the stopping condition is met. The algorithm returns with the best path found since its start ({\it Pbest}). The path construction favors the edges with high pheromone quantity, so the good paths are more likely to be chosen in the next iterations. For a better result, the search can be hybridized with local search procedures.

{\it ACO} is based on the ant algorithms designed to heuristically solve the {\it TSP}~\citep{dorigo2004ant}. The {\it ACO} pseudocode is sufficiently general to allow designing algorithms for a large set of problems ~\citep{dorigo2004ant}. The specificity is mainly observed in {\it UpdatePheromones} strategies and in {\it DaemonActions} method (it can miss).

\begin{verbatim}
procedure ACOMetaheuristic
    set parameters, initialize pheromone trails
    ScheduleActivities
        ConstructAntsSolutions
        UpdatePheromones
        DaemonActions       % optional
    end-ScheduleActivities
    return Pbest
end-procedure
\end{verbatim}

The {\it Ant Colony System} is an {\it ACO} version that starts from {\it Pbest}, an initial solution to the {\it TSP} instance and iteratively improves it. The initial solution can be randomly generated or other rapid construction method can be used. A constant number of $m$ ants form the ant colony that evolves and searches the shortest Hamiltonian tour in a complete weighted graph with $n$ vertices. The pheromone on edges form the matrix $(\tau_{ij})$ .

The initial pheromone quantity on each edge is constant. The three main activities in {\it ScheduleActivities} form an iteration and they can overlap. The solution construction is based on the pseudo-random-proportional rule, which is a combination of deterministic selection of the most promising vertex, and a probabilistic process of choosing the next vertex. The parameters that controls the solution constructions are: $\beta$, the weight of the heuristic function $\eta$ on edges (which is analytically expressed by the inverse of the distances) and $q_0$, the threshold for deterministic decisions.

If a specific ant stays in tertex $i$, the available vertices form the set $J_i$. The agent randomly generates a value $q\in [0,1]$. If $q\leq q_0$, then the next vertex $j$ is deterministically set using formula~(\ref{eq:c6}). Otherwise, $j$ is chosen using the transition probabilities from formula~(\ref{eq:c7}). After all the solutions are constructed, their lengths are computed and if a better solution was found, then {\it Pbest} is updated. Each ant updates the pheromone on its path using equation~(\ref{eq:c8}) where $\rho \in [0,1]$ is the evaporation parameter and $L_{initial}$ is the length of the initial tour. The current iteration ends by reinforcing the pheromone on the {\it Pbest} edges using the formula~\ref{eq:c9}, where $L_{best}$ is the length of {\it Pbest}. The iterations are repeated until a stopping criterion is met.

\begin{equation}\label{eq:c6}
j =arg max_{l\in J_i} {( \tau_{il} \cdot [\eta_{il} ]^\beta)}
\end{equation}

 \begin{equation}\label{eq:c7}
p_{ij} =
\frac{ \tau_{ij} \cdot [\eta_{ij} ]^\beta}
{\sum_{l\in J_i}\tau_{il} \cdot [ \eta_{il} ]^\beta }
\end{equation}

\begin{equation}\label{eq:c8}
\mbox{Local update:} \hspace{1cm}\tau_{ij}(t+1) = (1 - \rho) \cdot \tau_{ij}(t) + \rho  \frac{1}{n \cdot L_{initial}}
\end{equation}
\noindent where $L_{initial}$ is the length of the initial tour.

\begin{equation}\label{eq:c9}
\mbox{Global update:} \hspace{1cm}\tau_{ij}(t+1) = (1 - \rho) \cdot \tau_{ij}(t) + \rho  \frac{1}{ L_{best}}
\end{equation}
\noindent where $L_{best}$ is the length of the best tour.
Supplementary methods can be included in the iterations. It a current practice to use local improvement methods after each ant completes its tour. Such example of exchange heuristics are 2-Opt, 2.5-Opt or 3-Opt. In 2-Opt, a pair of edges is switched by other two edges that form a complete shorter tour~\citep{croes1958method}. The 2.5-Opt includes a better relocation of one vertex between two neighbor vertices~\citep{bentley1992fast}. The 3-Opt tries to optimize an existing tour by switching three edges~\citep{lin1965computer}.

{\it ACS} was further developed by many other algorithms. For example, the Inner Ant System introduces a new pheromone update rule for improving the ants’ local search~\citep{PinteaDumitrescu:2005:NewPheromone}.

Although initially designed as multi-agent systems that evolve independently, without central guidance, new ACO algorithms focus on
\begin{itemize}
\item exploiting more the information linked to good solutions or promising regions from the search space;
\item avoiding to be kept in a local optimum solution.
\end{itemize}

This guidance is designed as an activity that monitors the colony activity and reacts when it is needed. One such example is Foot Stepping technique described in~\citep{ventresca2004ant}. It is meant to decrease the pheromones on the best paths when stagnation manifests and thus to increase the exploration. A more complex strategy for avoiding the stagnation is described in~\citep{zhang2011guidance}. Adaptation is managed also by including pheromone alteration as  in~\citep{angus2005dynamic}. When a dynamic TSP with sudden modifications of the costs is approached, the pheromone values are modified. It is assumed that the algorithm knows when changes appear and immediately reacts.

All the guided actions are modules that automatically react when some pre-set criterion is met. To our knowledge, this paper is the first attempt to introduce totally external guidance to a working, solving process. Our investigation was founded on the good ability of humans to solve TSP instances. There are many studies in the literature that describe the process of trained or untrained TSP solving. For a good review see~\citep{MacGregorChu:2011:HumanPerformanceTSP}

The humans develop the ability to better solve an instance on repeated trials, they perform better on visually described instances with less vertices, they rarely provide crossing tours, and instinctively use the convex hull property (the vertices on the convex hull are traversed in their order). A visual interface for a game designed for assessing the human abilities for solving small TSP 2D instance is presented in~\citep{AcunaParada:2010:PeopleTSP}.

The pseudocode of Ant Colony Systems is illustrated in Algorithm \ref{alg:ACO}.

\begin{algorithm}
 \caption{Ant Colony System Algorithm\label{alg:ACO}}
 \DontPrintSemicolon

 \SetKwData{Left}{left}
 \SetKwData{This}{this}
 \SetKwData{Up}{up}
 \SetKwFunction{Union}{Union}
 \SetKwFunction{FindCompress}{FindCompress}
 \SetKwInOut{Input}{Input}
 \SetKwInOut{Output}{Output}

\Input{ProblemSize, $\emph{m}$, $\beta$,  $\rho$, $\sigma$, $\emph{q$_{0}$}$  }
\Output{$\emph{Pbest}$}
  $\emph{Pbest} \leftarrow $ CreateHeuristicSolution(ProblemSize);\;
  $\emph{Pbest$_{cost}$} \leftarrow $ Cost($\emph{Pbest}$);\;
  $\emph{Pheromone$_{init}$} \leftarrow$ $\frac{{1.0}}{ProblemSize\times Pbest_{cost}  }$;

  $Pheromone \leftarrow $ InitializePheromone($\emph{Pheromone$_{init}$}$);\;

 \SetAlgoLined
 \While{$\neg$StopCondition()}
  {
  \For{i $\emph{= 1 to}$ m}
  	{$\emph{S$_{i}$}$ $\leftarrow$ ConstructSolution(Pheromone, ProblemSize, $\beta$, $\emph{q$_{0}$}$);\;
  	$\emph{Si$_{cost}$}$ $\leftarrow$ Cost($\emph{S$_{i}$}$);\;
  	\If {Si$_{cost}$ $\leq$ $\emph{Pbest$_{cost}$}$}
  	{$\emph{Pbest$_{cost}$}$ $\leftarrow$ $\emph{Si$_{cost}$}$;\;
  	$\emph{Pbest} \leftarrow$ $\emph{S$_{i}$}$;
  	}
  	LocalUpdateAndDecayPheromone(Pheromone, $\emph{S$_{i}$}$, $\emph{Si$_{cost}$}$,  $\rho$);
 	}
 	GlobalUpdateAndDecayPheromone(Pheromone, $\emph{Pbest}$, $\emph{Pbest$_{cost}$}$, $\rho$);
  }
  \Return {Pbest};\;
\end{algorithm}

\section{Introducing new concepts based on Human Interaction}

As described in \citet{HolzingerEtAl:2016:iMLExperiment}, the human interaction is based on changing the ants' transition rules. In this paper we introduce two novel concepts: {\bf Human-Interaction-Matrix (HIM)} and {\bf Human-Impact-Factor (HIF)}.

\begin{itemize}
 \item[1.] {\bf Human-Interaction-Matrix (HIM)}. The creation of a HIM allows the human to control the ants actions. The user sets the probabilities for realizing his/her decisions. These values are dynamically interpreted by each ant, during its solution construction. This matrix may be dynamically modified by the user's decisions.

\item[2.] {\bf Human-Impact-Factor (HIF)}. The HIF is the variable interpretation of the HIM by each ant. When an ant is in node B, there are 50\% chances to be sent to C, based on HIM. In the other 50\% of the cases, when the ant does not go to C, it moves based on the algorithm' s transition rules, but C is excluded from the set of destinations. When an ant is in node C, there are 50\% chances to be sent in B and 10\% chances to be sent to E. If the ant does not go to B or E, then the implemented algorithm decides where to go, excluding B and E from the available nodes. All these interactions have no direct influence on the pheromones on the edges; they are just external guidance for the ants. If the user decides that one edge is no longer of interest, the corresponding HIM can be set to zero and the ants will not be influenced any more.
\end{itemize}

\noindent {\bf Pseudocode}. The new algorithm ACO-iML will include in the {\it ConstructionSolution} function the $HIF$ factor of probability.\\
{\it ConstructSolution(Pheromone, ProblemSize $\beta, q_0, HIF$)}

\begin{table*}[h]
\begin{tabular}{c}
\hspace{0.5cm}if ($q\leq HIF$) then $HIM$  identifies next move\\
\hspace{2.25cm}      else $ants$ identify next move\\
\end{tabular}
\end{table*}
\noindent where $q$ is an randomly generated number between $[0,1]$.

\begin{table}\centering
\begin{tabular}{c ccccc c cccccc}
  &A&B&C&D&E& &&A&B  &C  &D  &E  \\
A&0&0&0&0&0&  &A&N&0  &0  &0  &0  \\
B&0&0&0&0&0&  &B&0&N  &0.5&0  &0 \\
C&0&0&0&0&0&  &C&0&0.5&N  &0  &0  \\
D&0&0&0&0&0&  &D&0&0  &0  &N  &0  \\
E&0&0&0&0&0&  &E&0&0  &0.1&0  &N  \\
&&a)&&&&&&&b)
\end{tabular}\label{table:c1}

\caption{Example of Human-Interaction-Matrix (HIM) and Human-Impact-Factor
(HIF). Initially, no human intervention. The Human user sets 50\% chances to go from B to C and vice versa, and 10\% chances to go from C to E. In the other cases, the solver proceeds on its own. The loops are forbidden (N on the main diagonal).
}
\end{table}

\begin{figure}\centering
\includegraphics[scale=0.4]{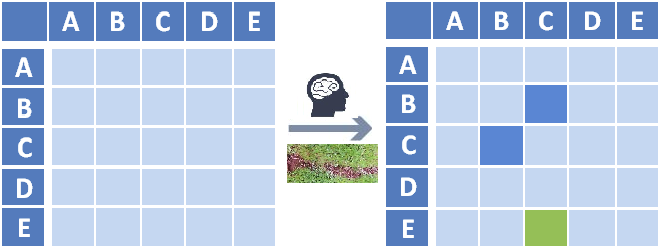}
\caption{Example of Human interaction in guiding ants, based on Table~\ref{table:c1}. Initially the matrix is empty; the {\it Human-Impact-Factor (HIF)} value HIF=0.5 (blue squares) force the ants to go from B to C in 50\% of the cases, and from C to B in 50\% of the cases. The {\it Human-Impact-Factor (HIF)} value HIF=0.1 (green square) forces the ants to go from C to E in 10\% of the cases. Note the symmetric set for the pair (B C) and the asymmetric set for (C E). In all the other cases, the ants move based on the transition rule only to the nodes having zero values (light blue squares). Loops excluded.}
\end{figure}

\section{Experimental Method, Setting and Results}

As an {\it aML} algorithm, ACO does not usually interact with the computing environment. The ants walk around and update the artificial pheromone on edges after each iteration. This procedure is repeated until a stopping criterion is met. Following the iML-approach, the human now can open the ACO black-box and can manipulate this algorithm by dynamically changing the behavior of the ants. This is done by introducing the transition rules (\ref{eq:c6}, \ref{eq:c7}) into the {\it else} path from the pseudocode in Section 2.


The current section illustrates several transition cases. As the current work considers asymmetric human settings, the HIM matrix can also be asymmetric. Starting from the matrix in Table 1 b), we shall describe one step of the algorithm evolution.

\begin{figure}
	\centering
		\includegraphics[scale=0.35]{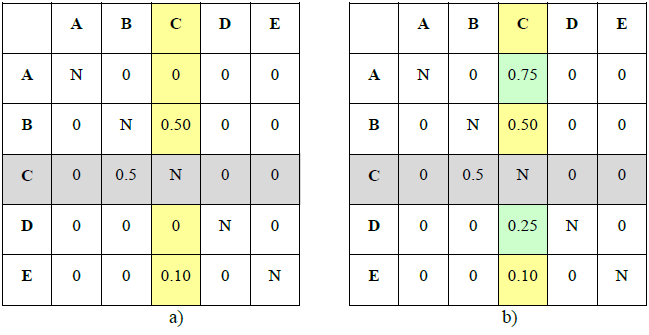}
	\label{fig:cer}
	\caption{a) Step 1: the ant is in C, line C is locked (gray background). b) Step 2, case 3: the ant ignores the human and normalizes the probabilities from rules (\ref{eq:c6}, \ref{eq:c7}) (green cells).}
\end{figure}

Let us suppose that an ant is initially dispatched in node C. In Fig.~\ref{fig:cer} a), the line C is locked (gray background), meaning that C is unavailable for further moves. Let us suppose that the transition rules (\ref{eq:c6}, \ref{eq:c7}) gave the following probabilities for moving {\it without human intervention} to available nodes: 30\% to A, 40\% to B, 10\% to D and 20\% to E.
The ant uniformly generates a random value $q_0$. We have the following cases, depending on which interval it falls:
\begin{itemize}
\item Case 1. If $q_0\in[0,0.5]$ then the ant follows the human decision and goes to node B.
\item Case 2. If $q_0\in[0.5,0.6]$ then the ant follows the human decision and goes to node E.
\item Case 3. If $q_0\in[0.6,1]$ then the ant follows the transition rules  (\ref{eq:c6}, \ref{eq:c7}). As in this case the nodes B and E are forbidden, the available nodes are A and D (zeroes in Fig. 4 a) and marked with green in Fig. 4 b)). As the transition without human intervention from C to A had the probability 30\%, the transition from C to D had to be made with 10\% probability, and only these two nodes are allowed, the ant normalizes these two values and computes new transition probabilities: 75\% for going to node A and 25\% for moving to D. The ant moves accordingly either to A or to D.
\end{itemize}

No matter where the ant just arrived, the corresponding line is locked and the process is repeated until the ant constructs a complete tour.

\subsection{Proposed iML Ant Colony Algorithm and Experimental Setup}

\begin{figure*}
	\centering
	\includegraphics[scale=0.35]{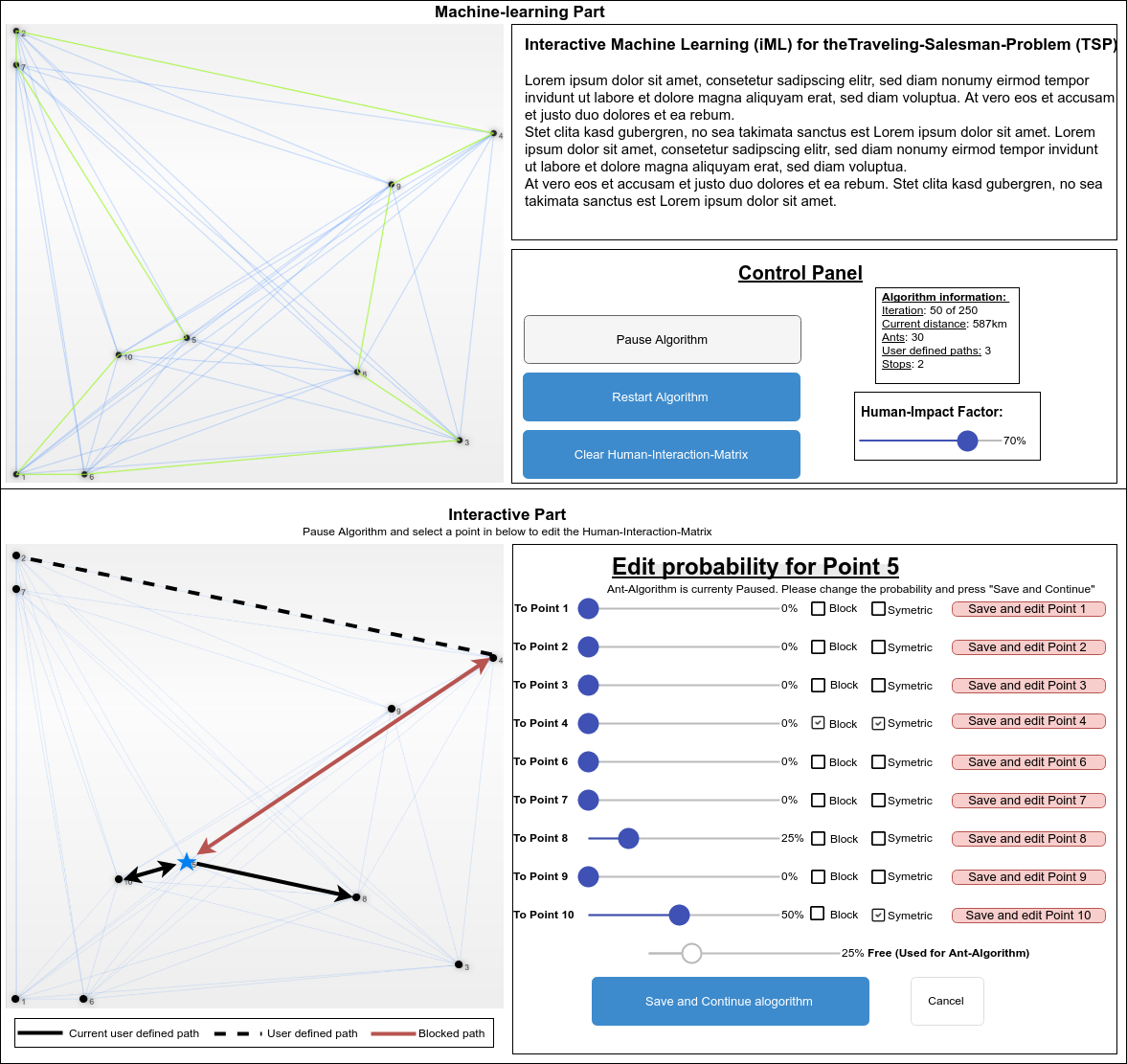}
	\caption{Draft of the GUI}
	\label{fig:GUI-Draft-iML}
\end{figure*}

In figure \ref{fig:GUI-Draft-iML} a draft of the GUI is presented. In the upper part of the machine-learning part is presented this contains all information from the standard Ant-Algorithm in pseudocode this can be written as~\ref{ACOiML}. The current shortest path is marked with a green path in different to the version described in ~\citep{HolzingerEtAl:2016:iMLExperiment} all lines have the same width. The HIF can be adjusted by a single slider, during the whole runtime of the algorithm.

The second part of the GUI allows the user to change the Human-Impact-Matrix. By clicking on one of the nodes the edit-mask of the Points opens. The mask now allows to adjust the probability for taking the track (as described above). For each Point "100\%" can be spread over the the other points. A path could also be locked completely by clicking the "Block" Button, if this is done, the rest of the pheromones won't be spread to this Points. All changes made by humans are summarized in a own picture (see \ref{fig:GUI-Draft-iML} bottom left).

So the work-flow can be summarized as :

$\longrightarrow$ Selection of the HIF

$\longrightarrow$ Start and initialization of the Ant-Algorithm

$\longrightarrow$ Pause of the Algorithm

$\longrightarrow$ Selection of Node

$\longrightarrow$ Modification of the HIM

$\longrightarrow$ Save the changed HIM

$\longrightarrow$ Another click on "Pause/Resume" continues the algorithm.

$\longrightarrow$ The steps above can be repeated as often as needed.

\begin{algorithm}[!t]
 \caption{Ant Colony Algorithm iML\label{ACOiML}}
 \DontPrintSemicolon

 \SetKwData{Left}{left}
 \SetKwData{This}{this}
 \SetKwData{Up}{up}
 \SetKwFunction{Union}{Union}
 \SetKwFunction{FindCompress}{FindCompress}
 \SetKwInOut{Input}{Input}
 \SetKwInOut{Output}{Output}

\Input{ProblemSize, $\emph{m}$, $\beta$,  $\rho$, $\sigma$, $\emph{q$_{0}$}$  }
\Output{$\emph{Pbest}$}
  $\emph{Pbest} \leftarrow $ CreateHeuristicSolution(ProblemSize);\;
  $\emph{Pbest$_{cost}$} \leftarrow $ Cost($\emph{Pbest}$);\;
  $\emph{Pheromone$_{init}$} \leftarrow$ $\frac{{1.0}}{ProblemSize\times Pbest_{cost}  }$;

  $Pheromone \leftarrow $ InitializePheromone($\emph{Pheromone$_{init}$}$);\;

 \SetAlgoLined
 \While{$\neg$StopCondition()}
  {
  \For{i $\emph{= 1 to}$ m}
  	{$\emph{S$_{i}$}$ $\leftarrow$ ConstructSolution(HIM, HIF, Pheromone, ProblemSize, $\beta$, $\emph{q$_{0}$}$);\;
  	$\emph{Si$_{cost}$}$ $\leftarrow$ Cost($\emph{S$_{i}$}$);\;
  	\If {Si$_{cost}$ $\leq$ $\emph{Pbest$_{cost}$}$}
  	{$\emph{Pbest$_{cost}$}$ $\leftarrow$ $\emph{Si$_{cost}$}$;\;
  	$\emph{Pbest} \leftarrow$ $\emph{S$_{i}$}$;
  	}
    LocalUpdateAndDecayPheromone(Pheromone, $\emph{S$_{i}$}$, $\emph{Si$_{cost}$}$,  $\rho$);
 	}
 	GlobalUpdateAndDecayPheromone(Pheromone, $\emph{Pbest}$, $\emph{Pbest$_{cost}$}$, $\rho$);\;
    \While {isUserInteraction()}
 	{
  		UpdateHumanInteractionMatrix(HIM);\;
  	}
  }
  \Return {P$_{best}$};\;
\end{algorithm}

The implementation is based on Java-Script. So it is a browser based solution which has the great benefit of platform independence and no installation is required.

For the test-setup we used 30 ants, 250 iterations. For the other parameters we choose fixed default values: $\alpha$ =1; $\beta$ =3; $\rho$ =0.1. The parameters are in the current prototype fixed, this makes it easier to compare the results.

After each iteration the current shortest path is calculated. If the new path is shorter than the old one, the green line will be updated.
For the evaluation on testing process there are some pre-defined data sets. From these data sets the best solution is known. The original data sets can be found on    \footnote{https://www.iwr.uni-heidelberg.de/groups/comopt/software/TSPLIB95/}.

The results of the pre-defined data sets can be compared with the optimal solution after finishing the algorithm by clicking on "Compare with optimal tour". The optimal tour is displayed with a red line.

\section{Discussion}

With the creation of the the Human-Impact-Factor and the Human-Interaction-Matrix the influence from the human was moved from Pheromone-Based influence to Ant-Decision based influence. This allows that the pheromones only were modified by the ants -- not by the human, so a possible flooding with pheromones can be avoided.


The current algorithm and the version from~\cite{HolzingerEtAl:2016:iMLExperiment} could be further extended as {\bf local search methods}. This is because it works better for a small number of nodes.

\begin{figure*}[ht!]
\centering
	\includegraphics[scale=0.7]{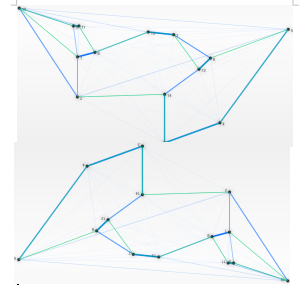}
	\caption{The iML human-in-the-loop approach as local search: A larger instance could be divided in clusters and each could be solved with our algorithm. In our case just a shorter way between the two graphs could be a solution of the problem.}
	\label{fig:both}
\end{figure*}

As an example we have included Figure~\ref{fig:both}, with the same {\it burma14.tsp} twice, so we have 28 nodes clustered. Making local search for each algorithms, the solution for the new instance with 28 nodes is based on the existing local solution adding the shortest distance between the two clusters.

\smallskip

 A larger instance should be clustered and the local results could be included to solve the larger instance. The increase of the cities is also possible and as in the preprocessing phase of subspace clustering~\citep{Hund2016,stoean2014support,matei2013optical}.

\section{Conclusion}\label{summary}
We demonstrated that the iML approach~\citep{Holzinger:2016:human-in-the-loop} can be used to improve current TSP solving methods. With this approach, we have enriched the way the human-computer interaction is used~\citep{AmershiEtAl:2014:iML}. This research is in-line with other successful approaches. Agile software paradigms, for example, are governed by rapid responses to change and continuous improvement. The agile rules are a good example on how the interaction between different teams can lead to valuable solutions. Gamification seeks for applying the Games Theory concepts and results to non-game contexts, for achieving specific goals. In this case, gamification could be helpful by considering the human and the computer as a coalition. There are numerous open research directions. The challenge now is to translate these approach to other similar problems, for example on protein folding, and at the same time to scale up on complex software applications.

%

\acks{We would like to thank our international colleagues and students who helped to test our algorithms on-line and provided valuable feedback.}


\end{document}